\documentclass[12pt]{conm-p-l}
\usepackage{amsmath}
\usepackage{amsfonts}
\usepackage{latexsym}
\usepackage{amssymb}
\usepackage[dvips]{graphics}

\newcommand{\R}{{\mathbf R}}
\newcommand{\C}{{\mathbf C}}

\newcommand{\kk}{{\mathbf k}}
\newcommand{\cat}{{\rm {cat }}}

\newcommand{\tc}{{\rm\bf {TC}}}
\newcommand{\supp}{\rm {supp}}

\newtheorem{theorem}{Theorem}[section]
\newtheorem{lemma}[theorem]{Lemma}
\newtheorem{corollary}[theorem]{Corollary}

\theoremstyle{definition}
\newtheorem{definition}[theorem]{Definition}
\newtheorem{example}[theorem]{Example}

\theoremstyle{remark}
\newtheorem{remark}[theorem]{Remark}

\numberwithin{equation}{section}

\title{Instabilities of Robot Motion}
\author{Michael Farber}

\address{School of Mathematical Sciences, 
Tel Aviv University, Ramat Aviv 69978, Israel}

\email{mfarber@tau.ac.il}

\thanks{Partially supported by a grant from the Israel Science Foundation and by the H. Minkowski Center for Geometry;
part of this work was done while the author visited  ETH in Z\"urich}
\date{\today}

\begin{document}

\begin{abstract}
Instabilities of robot motion are caused by topological reasons. In this paper we 
find a relation between the topological properties of a configuration space (the structure of its cohomology algebra)
and the character of instabilities, which are unavoidable in any motion planning algorithm.
More specifically, let $X$ denote the space of all admissible configurations of a mechanical system.
A {\it motion planner} is given by a splitting 
$X\times X = F_1\cup F_2\cup \dots \cup F_k$
(where $F_1, \dots, F_k$ are pairwise disjoint ENRs, see below) 
and by continuous maps
$s_j: F_j \to PX,$
such that $E\circ s_j =1_{F_j}$. 
Here $PX$ denotes the space of all continuous paths in $X$ (admissible motions of the system)
and $E: PX\to X\times X$
denotes the map which assigns to a path the pair of its initial -- end points.
Any motion planner determines an algorithm of motion planning for the system.
In this paper we apply methods of algebraic topology to study the minimal number of sets $F_j$
in any motion planner in $X$. We also introduce a new notion of {\it order of instability} of a motion planner; it describes 
the number of essentially distinct motions which may occur as a result of small perturbations of the input data.
We find  
the minimal order of instability, which may have motion planners on a given configuration space $X$. 
We study a number of specific problems: motion of a rigid body in $\R^3$, a robot arm, motion in $\R^3$ in the presence
of obstacles, and others.
\end{abstract}

\maketitle

\section{Motion Planning Problem}

In this article we will consider the problem of constructing a motion planning program for 
a large mechanical system. 
Such a program must function as follows: it should take as input pairs $(A, B)$ 
of admissible configurations of the system
and must produce as output, a description of a 
continuous motion of the system which starts at configuration $A$ and ends at configuration $B$.
Thus, after a motion planning program has been specified,
the movement of the system becomes a function of the input information $(A,B)$. 

A recent survey of algorithmic motion planning may be found in \cite{Sh}; see also textbook \cite{L}.

The goal of this paper is to study the character of discontinuities of the map 
\begin{eqnarray}
\qquad (A, B)\mapsto
\begin{array}{l}
 \mbox{continuous movement of the system}\\
\mbox{determined by}\, (A,B),
\end{array}\label{function}
\end{eqnarray}
which functionally emerge as instabilities of the robot motion.
We show that (\ref{function}) may be continuous only in very simple situations and hence instabilities 
of the robot motion are inevitable in most practically interesting cases. 
We will apply methods of algebraic topology (the cohomology theory)
to calculate the nature of the instabilities and to construct motion planning algorithms 
with a minimal order of instability or simply to show their existence.

Let $X$ be a metric space. 
We will regard points of $X$ as representing different configurations of a mechanical system. 
Usually, points of $X$ can be described 
by several parameters, which are 
subject to certain constraints (in the form of equations and inequalities).  
We will refer to $X$ as being our {\it configuration space}. 

We will always assume that $X$ is path connected, i.e. any pair of points $A, B\in X$ may be joined by a continuous path
$\gamma$ in $X$. This means that it is possible to bring our system, by a continuous movement,
from any given configuration $A$ to any given configuration
$B$. This assumption does not represent a restriction since in practical situations when the natural configuration space of a 
given system has several connected components, we may simply restrict our attention to one of them.

We will denote by $d(x,y)$ the metric 
(i.e. the distance function) in $X$. The metric $d$ itself will play
no significant role below, but the topology on $X$, determined by this metric, will be important to us. 

A continuous curve $\gamma: [0,1]\to X$ in $X$ describes a movement $\gamma(t)$, $0\leq t\leq 1$,
of the system starting at the 
initial position $A=\gamma(0)$ and ending at the final position $B=\gamma(1)$. 
We will denote by $PX$ the space of all continuous paths $\gamma: [0,1]\to X$. 
The path space $PX$ is a metric space (and hence, a topological space)
with respect to the metric 
\begin{eqnarray}
\rho(\gamma_1, \gamma_2) \, =\, \max_{t\in [0,1]}d(\gamma_1(t), \gamma_2(t)),\label{metric}
\end{eqnarray}
where $\gamma_1, \gamma_2\in PX$ are paths in $X$.

We will denote by 
\begin{eqnarray}
E: PX\to X\times X\label{fibration}
\end{eqnarray}
 the map which assigns to a path $\gamma\in PX$ 
the pair $(\gamma(0), \gamma(1))\in X\times X$ of initial -- final configurations. 
$E$ is a continuous map ({\it the endpoint map}).
Given a pair of configurations $(A,B)\in X\times X$, the preimage $E^{-1}(A,B)$ consists of all continuous paths $\gamma\in PX$,
which start
at $A$ and end at $B$. Therefore, the task of finding a continuous movement of the system from a configuration
$A$ to a configuration $B$ is equivalent to choosing an element of the set $E^{-1}(A,B)$. Since 
we assume that $X$ is path-connected,
the set $E^{-1}(A,B)$ is non-empty and so such a choice is always possible. 

A motion planning program is a rule specifying a continuous movement of the system once
the initial and the final configurations are given. Mathematically, this means that any motion planning program is 
a mapping 
\begin{eqnarray}
s: X\times X\to PX\label{section1}
\end{eqnarray}
from the space of all pairs of admissible configurations $X\times X$, 
to the space of all continuous movements of the system, $PX$, such that 
\begin{eqnarray}
E\circ s=1_{X\times X}. \label{section}
\end{eqnarray}
Here $1_{X\times X}: X\times X\to X\times X$ denotes the identity map and (\ref{section}) means precisely
that the path $s(A,B)$ assigned to a pair $(A,B)\in X\times X$, starts at the configuration $A$ and ends at the configuration $B$.

The first question to ask is the following: 

{\bf Question:} {\it Does there exist a continuous motion planning in $X$?} 

In other words, we ask
{\it whether it is possible to find a continuous map (\ref{section1}), satisfying (\ref{section})}. 

Using the language of the algebraic topology we may rephrase the above question as follows: 
the end-point map (\ref{fibration}) is a fibration (in the sense of Serre, see \cite{Sp});
any motion planning (\ref{section1}) has to be a section of $E$, and we ask if the fibration $E$ admits a continuous section.

Continuity of a motion planning strategy
$s$ means that for any small perturbation $(A',B')$ of the initial -- final pair of configurations $(A,B)$, 
the resulting movements of the system
$s(A',B'), \, s(A,B)\in PX$ are close to each other, with respect to the metric $\rho$, see (\ref{metric}).
Continuity of the motion planning program $s$ will guarantee that any small 
error in the description of the present position $A$ and the 
target position $B$ of the system will cause a small modification of the movement of the system, 
produced by the motion planner.

\begin{example}\label{ex1}
Suppose that we have to teach a robot, living on an island, how to move from any given position $A$ to any
given position $B$. Let us suppose first that the island has the shape of a convex planar domain $X\subset \R^2$. 
Then we may prescribe
the movement $s(A,B)$ from $A$ to $B$ in $X$ to be implemented along the straight line segment with a constant velocity.
This rule clearly defines a continuous motion planning $s: X\times X \to PX$.
\end{example}

\begin{example}\label{ex2} Suppose now that there is a lake in the middle of our island, 
and since our robot is not capable of swimming, it has to find its way over dry land.
It is easy to see that in this case there is no continuous motion planning strategy
$s: X\times X \to PX$ satisfying (\ref{section}). Indeed, suppose that such a continuous strategy $s$ exists. 
Fix two points $A$ and $B$ and consider the path $\gamma=s(A,B)$.
Now, suppose that point $A$ remains fixed but point $B$ starts moving and makes a circle $B_\tau$, where $0\leq \tau\leq 1$
around the lake, returning back to the initial position $B_0=B =B_1$.
Under this movement of point $B$ our motion planning program will produce a continuous curve 
$s(A, B_\tau)\in PX$ in the path space $PX$. 
We arrive at a contradiction since, on one hand, 
the final path $s(A,B_1)$ must be equal to the initial path $s(A,B)$, but on the other hand, it is homotopic (with endpoints fixed) to the product of the initial path $s(A,B_0)$ and the track of the point $B$, surrounding the lake. 

Hence we see that 
in Example \ref{ex2}, for any motion planning program $s: X\times X \to PX$, there always exists a pair $(A,B)\in X\times X$ of
initial -- final configurations, such that
$s$ is not continuous at $(A,B)$; this means that some arbitrarily close approximation $(A',B')$ of $(A,B)$ will 
cause a completely different movement $s(A',B')$ of the system. \end{example}

We will finish this section by citing the following result from \cite{F1}:

\begin{theorem}\label{contract} A globally defined continuous motion planning 
$s:X\times X \to PX$, $E\circ s=1_{X\times X}$, exists 
if and only if the configuration space $X$
is contractible.
\end{theorem} 

This explains why a continuous motion planning exists in Example \ref{ex1} and does not exist in Example \ref{ex2}.

\vskip 0.7cm

\section{Motion Planners}

In the following definition we describe the notion of a {\it motion planner in configuration space} $X$, 
which we will use in the rest of this paper.

\begin{definition}\label{planner}
Let $X$ be a path-connected topological space.
A motion planner in $X$ is given by finitely many subsets $F_1, \dots, F_k\subset X\times X$ and by continuous maps
$s_i: F_i\to PX$, where $i=1, \dots, k$, such that the following conditions are satisfied:
\begin{enumerate}
\item[(a)] the sets $F_1, \dots, F_k$ are pairwise disjoint $F_i\cap F_j=\emptyset$, $i\not= j$, and cover $X\times X$, i.e.
\begin{eqnarray}
X\times X = F_1\cup F_2\cup\dots \cup F_k; \label{splitting}
\end{eqnarray}
\item[(b)] $E\circ s_i =1_{F_i}$ for any $i=1, \dots, k$;
\item[(c)] each set $F_i$ is an ENR (see below).
\end{enumerate}
\end{definition}

We will refer to the subsets $F_i$ as to {\it local domains} of the motion planner.
The maps $s_i$ will be called {\it local rules} of the motion planner.

Condition (a) means that the sets $F_1, \dots, F_k$ partition the total space of all possible pairs $X\times X$.
Condition (b) requires that for any pair of configurations $(A, B)\in F_i$ the path $s_i(A,B)(t)$
is continuous as a function of the parameter $t\in [0,1]$, and $s_i(A,B)(0)=A$, 
$s_i(A,B)(1)=B$;
moreover, the path $s_i(A,B)(t)$ is a continuous function of the pair $(A,B)$ of initial -- final configurations as long as
the pair $(A,B)$ remains in the local domain $F_i$. 

By condition (c) we try to avoid pathological spaces.
Recall, a topological space 
$X$ is called a {\it Euclidean Neighbourhood Retract (ENR)} 
if it is homeomorphic to a subset of a Euclidean space $X'\subset \R^n$, such that
$X'$ is a retract of some open neighborhood $X'\subset U\subset \R^n$; in other words, $U\subset \R^n$ is open and 
there exists a continuous map $r: U\to X'$ such that $r(x)=x$ for all $x\in X'$. Such a continuous map $r$ is called a {\it retraction}.

The class of ENRs represents a reasonable class of topological spaces which (i) really appear as practically interesting
configuration spaces of mechanical systems, and (ii) possess important topological properties which allow 
considerable simplification of the theory.

Any motion planner determines a motion planning algorithm, as explained below.
Given a pair $(A, B)$ of initial - final configurations ({\it an input}), 
we determine the index $i\in \{1, 2, \dots, k\}$,
such that local domain $F_i$ contains $(A, B)$ (this index is unique); then we apply the local rule $s_i$ and 
produce the path $s_i(A,B)$ as an output.

In practical situations we are interested in constructing motion planners with the smallest possible number of local rules.
The configuration space $X$ depends on the nature of the mechanical 
system, which  we intend to control, and hence for us the space $X$ 
should be considered as given. Our decision consists of finding a motion planning strategy, i.e. in constructing 
a motion planner in a given topological space $X$. 

Hence we arrive at the following topological problems:

{\bf Problem 1}: {\it Given a topological space $X$, find (or estimate) 
the minimal number of local rules for a motion planner in $X$. }

{\bf Problem 2}: 
{\it Find practical ways of constructing a motion planner with the lowest possible number of local rules.}

\section{Example: Motion Planners on Polyhedra}

We will give here an explicit construction of a motion planner in $X$
assuming that $X$ is a connected finite-dimensional polyhedron.

Let $X^k$ denote the $k$-dimensional skeleton of $X$, i.e. the union of all simplices of $X$ of dimension $\leq k$.
The set $S_k=X^k - X^{k-1}$ is the union of interiors of all $k$-dimensional simplices.
Here $k=0, 1, \dots, n$, where $n=\dim X$ denotes the dimension of $X$.
Denote
$$F_i =\bigcup_{k+\ell=i} S_k\times S_\ell\subset X\times X, \quad \mbox{where}\quad i=0, 1, \dots, 2n.$$
Each set $F_i$ is an ENR (since it is homeomorphic to a disjoint union of balls), $F_i$ and $F_j$ are disjoint for $i\not=j$, 
and the union $F_0\cup F_1\cup \dots \cup F_{2n}$
equals $X\times X$.

We will describe a continuous local rule $s_i: F_i\to PX$ for each $i=0, 1, \dots, 2n$.
The set 
$F_i$ is the union of disjoint sets $S_k\times S_\ell$, $k+\ell =i$, which are both closed and open in $F_i$. Hence it is enough to 
construct a continuous map $s_i: S_k\times S_\ell\to PX$, where $i=k+\ell$ and $E\circ s_i=1$.

Fix a point in the interior of each simplex of $X$; we will refer to this point as to the {\it center} of the simplex.
For any ordered pair of simplices fix a continuous path connecting the centers of the simplices.
Now, given a pair of points $(A, B)\in S_k\times S_\ell$, we will set $s_i(A,B)$ as the path in $X$ which first 
goes along the straight line segment connecting $A$ with the center of the simplex containing $A$, then along the precomputed 
path from the center of the simplex containing $A$ to the center of the simplex containing $B$, and finally going to $B$
along the straight line segment.

\begin{corollary}\label{polyhedron}
If $X$ is an $n$-dimensional polyhedron then it admits a motion planner with $2n+1$ local rules. 
In particular, any graph admits a motion planner with three local rules.
\end{corollary}

\section{Order of Instability of a Motion Planner}

Besides the total number of local rules, the motion planners could be characterized by their orders of instability:

\begin{definition}\label{unsable} The order of instability of a motion planner (see Definition \ref{planner})
at a pair of initial -- final configurations $(A,B)\in X\times X$ is defined as the largest number $r$ such that 
any neighborhood of $(A,B)$ has a nontrivial intersection with $r$ distinct local domains among $F_1, \dots, F_k$.
\end{definition}

In other words, the order of instability of a motion planner at a pair $(A,B)\in X\times X$ is defined as the largest $r$, such that
$(A,B)$ belongs to intersection 
\begin{eqnarray*}
\bar F_{i_1}\cap \bar F_{i_2}\cap \dots \cap \bar F_{i_r},\quad\mbox{where}\quad 1\leq i_1<i_2<\dots< i_r\leq k.
\end{eqnarray*} 

If $(A',B')\in X\times X$ is a small perturbation of $(A,B)$, it may lie in one of the $r$ local domains
$F_{i_1},  F_{i_2}, \dots, F_{i_r}$.

\begin{definition}
The order of instability of a motion planner is defined as the maximum of the orders of instability at all possible pairs
$(A,B)\in X\times X$. Equivalently, the order of instability of a motion planner is the largest $r$ such that the closures of some $r$
among the local domains $F_1, \dots, F_k$ have a non-empty intersection: 
\begin{eqnarray*}
\bar F_{i_1}\cap \bar F_{i_2}\cap \dots \cap \bar F_{i_r}\, \not=\, \emptyset
\quad\mbox{where}\quad 1\leq i_1<i_2<\dots< i_r\leq k.
\end{eqnarray*} 
\end{definition}

Clearly, the order of instability of a motion planner does not exceed the total number of local rules, i.e. 
\begin{eqnarray}
1\leq r\leq k.\label{degree}
\end{eqnarray} 
Another remark: the order of instability equals one, $r=1$, if and only if $k=1$, i.e. the local rules 
produce a continuous globally defined motion planning $s:X\times X\to PX$; as we know from Theorem \ref{contract}, this may happen only when the configuration space $X$ is contractible.

\setlength{\unitlength}{1cm}
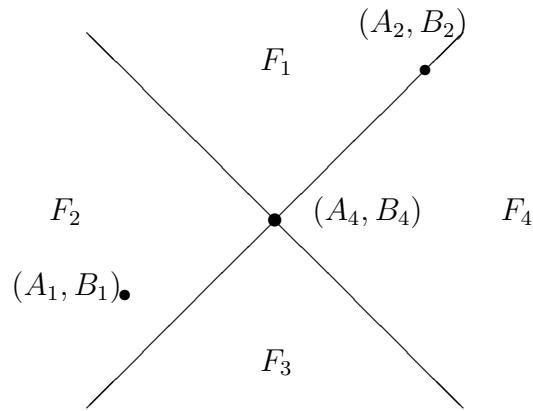
\begin{figure}[h]
  \begin{center}
\begin{picture}(10,6)
\linethickness{0.3mm}

\put(5,3){\circle*{0.18}}
\put(5,3){\line(1,1){2.5}}
\put(5,3){\line(1,-1){2.5}}
\put(5,3){\line(-1,1){2.5}}
\put(5,3){\line(-1,-1){2.5}}
\put(7,5){\circle*{0.15}}
\put(6.1,5.5){$(A_2,B_2)$}
\put(4.8, 5){$F_1$}
\put(4.8, 1){$F_3$}
\put(2, 3){$F_2$}
\put(8, 3){$F_4$}
\put(3,2){\circle*{0.15}}
\put(1.5,2){$(A_1,B_1)$}
\put(5.5,3){$(A_4,B_4)$}
\end{picture}
\end{center} \caption{Motion planner with local domains $F_1, F_2, F_3, F_4$.
The order of instability at any pair $(A_i,B_i)$ equals $i$, where $i=1, 2, 4$}\label{fig3}

\end{figure}

The order of instability represents a very important functional characteristic of a motion planner.
If the order of instability equals $r$ then 
there exists
a pair of initial - final configurations $(A,B)\in F_j$ such that arbitrarily close to $(A,B)$ 
there are $r-1$ pairs of configurations
$(A_1, B_1)$, $(A_2, B_2)$, $\dots$,
$(A_{r-1}, B_{r-1})$ (which are different small perturbations of $(A,B)$), belonging to distinct sets $F_i$, where $i\not=j$.
This means that small perturbations of the input data $(A,B)$ may lead to $r$ essentially distinct motions suggested by the motion
planning algorithm. 

On the other hand, if the order of instability equals $r$, 
there are no input data $(A,B)$ such that their small perturbations may have more than $r$
essentially distinct motions.
In practical situations we are interested in motion planners with a degree of instability as low as possible.

{\bf Problem 3}: {\it Given a path-connected topological space $X$, 
find (or estimate) the minimal order of instability that may have a motion planner in $X$.
Find (describe) motion planners in $X$ with the minimal order of instability.}

Clearly, there may exist motion planners with a low order of instability and a large number of local rules. 
However, as we shall see below, the order of instability coincides with the number of local rules, assuming that 
the number of local rules $k$ is minimal.

\section{Invariant $\tc(X)$}
 
In paper \cite{F1} we introduced invariant $\tc(X)$, which measures the topological complexity of the motion planning problem 
in $X$. Invariant $\tc(X)$ allows us to answer Problems 1 and 3 raised above. For convenience of the reader 
we will give here the definition and will briefly review the basic properties of $\tc(X)$.

\begin{definition} {\rm (See \cite{F1}.) }
Given a path-connected topological space $X$,  the topological complexity of motion planning
in $X$ is defined as the minimal number ${\rm \bf {TC}}(X)=k$, such that the Cartesian product 
$X\times X$ can be covered by $k$ open subsets
\begin{eqnarray}\label{cover}
X\times X = U_1\cup U_2\cup \dots \cup U_k,
\end{eqnarray}
where for any $i=1, 2, \dots, k$ there exists a continuous map 
\begin{eqnarray}
s_i: U_i\to PX\quad \mbox{with}\quad E\circ s_i =1_{U_i}.\label{local}
\end{eqnarray}
If no such $k$ exists, we will set $\tc(X)=\infty$.
\end{definition}

In \cite{F1} we proved that $\tc(X)$ {\it is a homotopy invariant of $X$, i.e. $\tc(X)$ depends only on the homotopy type of} $X$.

For example, $\tc(X)=\tc(Y)$ where $X=S^1$ is a circle, and $Y=\C-\{0\}$ is a punctured plane.

In paper \cite{F1}  we gave an estimate for $\tc(X)$ from below in terms of the cohomology algebra of $X$. 
The lower bound provides topological restrictions on the number of open sets $U_i$ in any open cover (\ref{cover}).
For example, in the case when $X$ is the 2-dimensional sphere $S^2$, any cover (\ref{cover}) must have at least
three open sets.

Also, according to 
\cite{F1},
$\tc(X)$ has an upper bound in terms of the dimension of $X$, namely
\begin{eqnarray}
\tc(X) \leq 2\dim(X)+1,\label{upper}
\end{eqnarray}

The meaning of the upper bound, compared with the lower bound, is completely
different: there always exists an open cover (\ref{cover}) with $2\cdot\dim(X)+1$ open sets $U_i$ and continuous motion planning
programs $s_i:U_i\to PX$. 

Now we will give an improvement of (\ref{upper}).

Recall that a topological space $X$ is called $r$-connected if for any $i\leq r$ 
any continuous map $S^i\to X$ of a sphere of dimension 
$i$ into $X$ can be extended to a 
continuous map of a ball $D^{i+1}\to X$. 
Examples: a path-connected space is $0$-connected, a simply-connected space is $1$-connected. 

\begin{theorem}\label{upper2} Let $X$ be an $r$-connected CW polyhedron. Then 
\begin{eqnarray}
\tc(X) \, < \, \frac{2\cdot \dim(X)+1}{r+1}+1.\label{upper1}
\end{eqnarray}
\end{theorem}

\begin{proof}
Theorem \ref{upper2} follows directly from Theorem 5 of paper of A. S. Schwarz \cite{Sz}, 
where the notion of a genus of a fiber space
was introduced. The topological complexity $\tc(X)$ can be viewed as the
genus of the path space fibration $E: PX\to X\times X$, which has the base of dimension $\dim(X\times X)=2\dim X$. 
The fiber is
homotopy equivalent to the space $\Omega X$ of based loops in $X$. If $X$ is $r$-connected then $\Omega X$ is $(r-1)$-connected, i.e. it is aspherical in dimensions $< r$. 
Inequality (\ref{upper1}) now follows applying Theorem 5 from \cite{Sz}.
\end{proof}

\begin{corollary}\label{simple}
Let $X$ be a simply connected polyhedron.
Then 
\begin{eqnarray}
\tc(X)\leq \dim X +1.
\end{eqnarray}
\end{corollary}
\begin{proof} Theorem \ref{upper2} applies with $r=1$ and gives $\tc(X) < \dim(X) +1+\frac{1}{2}$, which is equivalent
to our statement. 
\end{proof}
\section{Order of Instability and $\tc(X)$}

The next result gives a partial answer to Problems 
1 and 3, see above. 

\begin{theorem}\label{coincide} Let $X$ be a connected $C^\infty$-smooth manifold. Then: 
(1) the minimal integer $k$, such that $X$ admits a motion planner (in the sense of Definition \ref{planner})
with $k$ local rules, 
equals $\tc(X)$.  Moreover,
(2) the minimal integer $r>0$, such that $X$ admits a motion planner with order of instability $r$, equals $\tc(X)$.
\end{theorem}

We may restate this Theorem as follows:

\begin{theorem}\label{coincide1} 
Let $X$ be a connected smooth manifold. 
Then for any motion planner in $X$, the number of local rules $k$
and the order of instability $r$ are at least $\tc(X)$, i.e.
$k\geq \tc(X),$  $r\geq \tc(X).$
Moreover, there exists a motion planner in $X$ with $k=\tc(X)$ local rules and with order of instability
$r=\tc(X)$.
\end{theorem}

In this section we will prove the following statement which is the main ingredient in the proof of Theorem \ref{coincide}:

\begin{theorem}\label{cor1} Suppose that $X$ is a connected smooth manifold. Let 
$$X\times X= F_1\cup F_2\cup\dots\cup F_k,\quad 
s_1, \dots, s_k: F_i\to PX,$$
be a motion planner in $X$ with the minimal number of local rules, $k=\tc(X)$. Then the intersection 
of the closures of the local domains
\begin{eqnarray}
\bar F_1\cap \bar F_2\cap \dots \cap \bar F_k\, \not= \emptyset\label{intersection1}
\end{eqnarray} 
is not empty and thus the order of instability of this motion planner equals $\tc(X)$. 
\end{theorem}

\begin{remark} In Theorems \ref{coincide}, \ref{coincide1} and \ref{cor1} we assume that the configuration space $X$
is a smooth manifold. We use this assumption in the proof since we apply smooth partitions of unity and 
Sard's Theorem. A different piecewise linear
technique could be used instead. One may show that Theorems \ref{coincide}, \ref{coincide1} and \ref{cor1} 
hold assuming only that $X$ is a polyhedron.
\end{remark}

\begin{proof}[Proof of statement (1) of Theorem \ref{coincide}.] 
Suppose that $X$ admits a motion planner in the sense of Definition \ref{planner}
with $k$ local domains $F_1, \dots, F_k\subset X\times X$ and with the corresponding local rules
$s_i: F_i\to PX$, where $i=1, \dots, k$. Let us show that then $k\geq \tc(X)$. This claim would follow once we know that 
{\it one may enlarge the local domains $F_i$ to open sets $U_i$ such that over each $U_i$
there exists a continuous motion planning map} (\ref{local}).

We will use the next well-known property of the ENRs: {\it
If $F\subset X$ and both spaces $F$ and $X$ are ENRs then there is an open neighborhood $U\subset X$ of $F$ and a retraction $r: U\to F$ such that the inclusion $j:U\to X$ is homotopic to $i\circ r$, where $i: F\to X$ denotes the inclusion.} See \cite{D}, 
chapter 4, \S 8 for a proof.

Using the fact that both 
$F_i$ and $X\times X$ are ENRs, we find that there exists an open 
neighborhood $U_i\subset X\times X$ of the set
$F_i$ and a continuous homotopy 
$h^i_\tau: U_i\to X\times X$, where $\tau\in [0,1]$, such that 
$h^i_0: U_i\to X\times X$
is the inclusion and $h^i_1$ is a retraction
of $U_i$ onto $F_i$. We will describe now a continuous map $s'_i: U_i\to PX$ with $E\circ s'_i=1_{U_i}.$
Given a pair $(A,B)\in U_i$, the path $h^i_\tau(A,B)$ in $X\times X$ is a pair of paths $(\gamma, \delta)$, where
$\gamma$ is a path in $X$ starting at the point $\gamma(0)=A$ and ending at a point $\gamma(1)$, and $\delta$ is a path in $X$
starting at $B=\delta(0)$ and ending at $\delta(1)$. Note that the pair $(\gamma(1), \delta(1))$ belongs to $F_i$; therefore the motion 
planner $s_i: F_i\to PX$ defines a path 
$$\xi=s_i(\gamma(1),\delta(1))\, \in \, PX$$ 
connecting the points $\gamma(1)$ and $\delta(1)$. 
Now we set $s'_i(A,B)$ to be the concatenation of $\gamma$, $\xi$, and $\delta^{-1}$ (the reverse path of $\delta$):
$$s'_i(A,B) \, =\, \gamma\cdot\xi\cdot\delta^{-1}.$$
Formally, $s'_i(A,B)$ is given by the formula
\begin{eqnarray*}
s'_i(A,B)(t) \, =\, \left\{
\begin{array}{lll}
\gamma(3t)&\mbox{for}& 0\leq t\leq 1/3,\\ \\
\xi(3t-1)&\mbox{for}&1/3\leq t\leq 2/3,\\ \\
\delta(3-3t)&\mbox{for}&2/3\leq t\leq 1.
\end{array}
\right.
\end{eqnarray*}
Continuity of $s'_i(A,B)(t)$ as a function of $A, B, t$ is clear. This proves the italicized claim.

Now we want to show that $X$ always 
admits a motion planner (in the sense of Definition \ref{planner}) with the number of local domains
equal to $k=\tc(X)$. Let 
\begin{eqnarray}
U_1\cup U_2\cup \dots \cup U_k=X\times X,\quad\mbox{where}\quad k=\tc(X),\label{cover1}
\end{eqnarray}
be an open cover such that for any $i=1, \dots, k$ there exists a continuous motion planning map $s_i:U_i\to PX$
with $E\circ s_i=1_{U_i}$. Find a smooth partition of unity $\{f_1, \dots, f_k\}$ subordinate to the
cover (\ref{cover1}). Here $f_i: X\times X\to [0,1]$ are smooth functions, $i=1, \dots, k$, with the support of $f_i$ being a subset of
$U_i$, and such that for any pair $(A,B)\in X\times X$, it holds that
$$f_1(A,B)+ f_2(A,B)+\dots+f_k(A,B)=1.$$
Recall that the support $\supp(f)$ of a continuous function $f:X\times X\to \R$
is defined as the closure of the set $\{(A,B)\in X\times X; f(A,B)\neq 0\}$. 

Choose numbers $0<c_i<1$, where $i=1, \dots, k$, with $c_1+\dots +c_k=1$,
such that each $c_i$ is a regular value of the function $f_i$. Such numbers exist by the Sard's Theorem.
Let a subset $V_i\subset X\times X$, where $i=1, \dots, k$, be defined by the following system of inequalities
\begin{eqnarray*}
\left\{
\begin{array}{ll}
f_j(A,B) < c_j&\mbox{for all}\quad j<i,\\ \\
f_i(A,B)\geq c_i.&
\end{array}
\right.\label{fine}
\end{eqnarray*}
One easily checks that:
\begin{enumerate}
\item[(a)] each $V_i$ is a manifold with boundary and hence an ENR;

\item[(b)] $V_i$ is contained in $U_i$; therefore, the local rule $s_i:U_i\to PX$ restricts onto $V_i$ and defines a local rule 
over $V_i$;

\item[(c)] the sets $V_i$ are pairwise disjoint, $V_i\cap V_j=\emptyset$ for $i\not= j$;

\item[(d)] $V_1\cup V_2\cup \dots \cup V_k = X\times X$. 
\end{enumerate}
Hence we see that the submanifolds $V_i$ and the local rules $s_i|_{V_i}$ define a motion planner in the sense of Definition 
\ref{planner} with $\tc(X)$ local domains. 

This completes the proof of statement (1) of Theorem \ref{coincide}. \end{proof}

The proof of statement (2) of Theorem \ref{coincide} uses the following Lemma.

\begin{lemma}\label{intersection} 
Let $X$ be a path-connected metric space.
Consider an open cover 
$$X\times X= U_1\cup U_2\cup \dots\cup U_k$$
such that for any $i=1, \dots, k$ there exists a continuous map $s_i: U_i\to PX$ with $E\circ s_i = 1_{U_i}$. 
Suppose that for some integer $r$ any intersection 
$$U_{i_1}\cap U_{i_2}\cap \dots \cap U_{i_r} \, =\, \emptyset$$ 
is empty, where $1\leq i_1< i_2< \dots<i_r$. Then $\tc(X)<r$, i.e.
there exists an open cover 
$$X\times X= W_1\cup W_2\cup \dots\cup W_{r-1},$$ 
consisting of $r-1$ open sets $W_i$, and continuous maps 
$s'_i: W_i\to PX$, where $i=1, \dots, r-1$,
such that $E\circ s'_i=1_{W_i}$. 
\end{lemma}
\begin{proof}  Let $f_i: X\times X\to [0,1]$, where $i=1, \dots, k$, be a partition of unity
subordinate to the cover $\{U_1, \dots , U_k\}$. This means that each $f_i$ is a continuous function, the support of $f_i$ is
contained in the set $U_i$, and 
$$f_1(A,B)+\dots+ f_k(A,B) =1$$
for any $A,B\in X$. 
Here we use the fact that $X\times X$ is a metric
space and hence for any of its open covers there exists a subordinate partition of unity, see \cite{K}.

For any nonempty subset $S\subset \{1, \dots, k\}$ let 
$$W(S)\subset X\times X$$ 
denote the set of all
pairs $(A,B)\in X\times X$, such that for any $i\in S$ it holds that $f_i(A,B)>0$, and for any $i'\notin S$,
$$f_i(A,B) > f_{i'}(A,B).$$

One easily checks that:

{\it (a) each set $W(S)\subset  X\times X$ is open;

(b) $W(S)$ and $W(S')$ are 
disjoint if neither $S\subset S'$ nor $S'\subset S$;

(c) if $i\in S$, then $W(S)$ is contained in $U_i$; therefore there exists a continuous motion planning 
over each $W(S)$;

(d) the sets $W(S)$ with all possible nonempty $S$ such that $|S|<r$, form a cover of $X\times X$.}

To prove (d), suppose that $(A,B)\in X\times X$. Let $S$ be the set of all indices 
$i\in \{1, \dots, k\}$,
such that $f_i(A,B)$ equals the maximum of $f_j(A,B)$, where $j=1, 2, \dots, k$. 
Then clearly the pair $(A,B)$ belongs to $W(S)$.
The pair $(A,B)$ lies in the intersection of the sets $U_j$ with $j$ in $S$. Since we assume that the intersection of any $r$ sets
$U_1, U_2, \dots, U_k$ is empty, we conclude that $|S|<r$.

Let $W_j \subset X\times X$ 
denote the union of all sets $W(S)$, where $|S|=j$. Here $j=1, 2, \dots, r-1.$
The sets $W_1, \dots, W_{r-1}$ form an open cover of $X\times X$. 
If $|S|=|S'|,$ then the corresponding sets $W(S)$ and $W(S')$ either coincide (if $S=S'$), or are disjoint. 
Hence we see (using (c)) that there exists a continuous motion planning over each open set $W_j$. 

This completes the proof. 
\end{proof}

\begin{proof}[Proof of statement (2) of Theorem \ref{coincide}.] Any motion planner with $\tc(X)$ local rules will have degree of instability
$r\leq \tc(X)$, see (\ref{degree}). Hence to prove statement (2) it is enough to show that the degree of instability of any
motion planner in $X$ satisfies 
$r\geq \tc(X).$ 
Suppose that $F_1, \dots, F_k\subset X\times X$, $s_i: F_i \to PX$  is a motion planner
with degree of instability $r$. Then any intersection of the form
\begin{eqnarray}
\bar F_{i_1}\cap \dots \cap \bar F_{i_{r+1}} \, =\, \emptyset, \label{empty}
\end{eqnarray}
is empty, where $1\leq i_1< i_2< \dots < i_{r+1}\leq k$. For any index $i=1, \dots, k$ fix a continuous function
$f_i: X\times X\to [0,1]$, such that $f_i(A,B)=1$ if and only if pair $(A,B)$ belongs to $\bar F_i$. Let $\phi: X\times X\to \R$
be the maximum of (finitely many) functions of the form
$f_{i_1}+ f_{i_2}+\dots+f_{i_{r+1}}$ for all increasing sequences $1\leq i_1< i_2< \dots < i_{r+1}\leq k$ of length $r+1$.
We have: 
$$\phi(A,B)<k$$ 
for any pair $(A,B)\in X\times X$, as follows from (\ref{empty}). 

Let $U_i\subset X\times X$ denote the set of all $(A,B)$ such that 
$$k\cdot f_i(A,B) > \phi(A,B).$$ 
Then $U_i$ is open
and contains $\bar F_i$, and hence the sets $U_1, \dots, U_k$
form an open cover of $X\times X$. On the other hand, any intersection
$$U_{i_1}\cap U_{i_2}\cap \dots \cap U_{i_{r+1}} =\emptyset$$
is empty.

As in the proof of statement (1), we may assume that $U_1, \dots, U_{k}$ are small enough so that over each
$U_i$ there exists a continuous motion planning (here we use the assumption that each $F_i$ is an ENR). 
Applying Lemma \ref{intersection} we conclude that $\tc(X)\leq r$. 
This completes the proof. \end{proof}

\section{A Cohomological Lower Bound for $\tc(X)$}

We will briefly recall a result from \cite{F1} giving a lower bound on $\tc(X)$. 
It is quite useful since it allows an effective computation of $\tc(X)$ in many examples.

Let $\kk$ be a field; one may always assume that $\kk = \R$ is the field of real numbers. 
The singular cohomology $H^\ast(X;\kk)$ is a graded $\kk$-algebra with the multiplication
\begin{eqnarray}
\cup: H^\ast(X;\kk)\otimes H^\ast(X;\kk)\to H^\ast(X;\kk)\label{prod}
\end{eqnarray}
given by the cup-product, see \cite{DNF}, \cite{Sp}. 
The tensor product $H^\ast(X;\kk)\otimes H^\ast(X;\kk)$ is also a graded $\kk$-algebra
with the multiplication 
\begin{eqnarray}\label{signs}
(u_1\otimes v_1)\cdot (u_2\otimes v_2) = (-1)^{|v_1|\cdot |u_2|}\, u_1u_2\otimes v_1v_2.
\end{eqnarray}
Here $|v_1|$ and $|u_2|$ denote the degrees of cohomology classes $v_1$ and $u_2$ correspondingly.
The cup-product (\ref{prod}) is an algebra homomorphism.

\begin{definition} The kernel of homomorphism (\ref{prod}) is called {\it the ideal of the zero-divisors} of $H^\ast(X;\kk)$.
The {\it zero-divisors-cup-length} of $H^\ast(X;\kk)$ is the length of the longest nontrivial product in the ideal of the zero-divisors
of $H^\ast(X;\kk)$.
\end{definition}

\begin{theorem}\label{lower} The number $\tc(X)$ 
is greater than the zero-divisors-cup-length of 
$H^\ast(X;\kk)$.
\end{theorem}

See \cite{F1} for a proof. 

We will illustrate Theorem \ref{lower} by calculating $\tc(X)$, when $X$ is a graph.

\begin{theorem}\label{graph1}
Let $X$ be a connected graph. 
Then 
\begin{eqnarray}\label{graph}
\tc(X) = \left\{
\begin{array}{lll} 
1, & \mbox{if} & b_1(X)=0,\\
2, & \mbox{if} & b_1(X)=1,\\
3, & \mbox{if} & b_1(X)\geq 2.
\end{array}
\right.
\end{eqnarray}
Here $b_1(X)$ denotes the first Betti number of $X$.
\end{theorem}

\begin{proof}
To prove (\ref{graph}) we first note that $\tc(X)\leq 3$ by Corollary \ref{polyhedron}.
Also, we know that $\tc(X)>0$ if $X$ is not contractible, i.e. if $b_1(X)>0$. 

Let us show 
(using Theorem \ref{lower})
that $\tc(X)\geq 3$ for
$b_1(X)\geq 2$. Indeed, taking $\kk=\R$ we find that there are two linearly independent classes $u_1, u_2\in H^1(X;\R)$
and in the tensor product algebra $H^\ast(X;\R)\otimes H^\ast(X;\R)$ the product
$$(1\otimes u_1-u_1\otimes 1)\cdot (1\otimes u_2-u_2\otimes 1)= u_2\otimes u_1- u_1\otimes u_2 \not=0$$
is nontrivial. Hence by Theorem \ref{lower}, we find $\tc(X)\geq 3$ and hence we obtain $\tc(X)=3$.

We are left with the case $b_1(X)=1$. Then $X$ is homotopy equivalent to the circle and therefore, using homotopy invariance
of $\tc(X)$, we have $\tc(X)=\tc(S^1)$. One may easily construct a motion planner on the circle $S^1$ with two local rules; hence
$\tc(X)=2$. 
\end{proof}

\section{Rigid Body Motion in $\R^3$}

Let $SE(3)$ denote the group of all orientation-preserving isometric transformations $\R^3\to \R^3$.
Points of $SE(3)$ describe movements of a rigid body in the 3-dimensional space $\R^3$.
The dimension of $SE(3)$ equals 6. Any orientation-preserving isometric transformation $\R^3\to \R^3$
can be written in the form $x\mapsto Ax+b$, where $b\in \R^3$ and $A\in SO(3)$ is an orthogonal matrix. 

\begin{theorem}\label{body} The topological complexity of $SE(3)$ equals 4. Therefore, any motion planner having 
$SE(3)$ as the configuration space (for example, any motion planner moving a rigid body in $\R^3$), 
will have points with order of instability $\geq 4$. Moreover, there exists 
a motion planner on $SE(3)$, having order of instability $4$, i.e. having no points of instability of order greater than $4$.
\end{theorem}

Proof of Theorem \ref{body} will use the following Lemma, suggested by S. Weinberger:

\begin{lemma}\label{shmuel} Let $G$ be a connected Lie group. Then 
\begin{eqnarray}
\tc(G)=\cat(G).\label{cat}
\end{eqnarray}
\end{lemma}
Here we use the following notation:
$\cat(X)$ denotes the Lusternik - Schnirelman category of a topological space $X$, which is defined as 
the minimal integer $k$, such that
$X$ admits an open cover $U_1\cup U_2\cup \dots \cup U_k=X$, such that each inclusion $U_i\to X$ is 
homotopic to a constant map. 
We refer to \cite{J} for more information. 

In general, there is the following relation between the topological complexity
$\tc(X)$ and the Lusternik - Schnirelman category $\cat(X)$, see \cite{F1}, formula (4):
\begin{eqnarray}
\cat(X)\leq \tc(X)\leq \cat(X\times X).\label{between}
\end{eqnarray}
Lemma \ref{shmuel}
claims that left inequality in (\ref{between}) is an equality if $X=G$ is a connected Lie group.
It is not true that left inequality in (\ref{between}) is an equality for all topological spaces $X$;
for example, we know that for the sphere $X=S^n$ with $n$ even, 
 $\tc(S^n)=3$ while $\cat(S^n)=2$, see \cite{F1}.

\begin{proof}[Proof of Lemma \ref{shmuel}]: 
Assume that $\cat(G)\leq k$, i.e. we may find an open cover $G=U_1\cup U_2\cup \dots \cup U_k$ such that each inclusion $U_i\to G$ is null-homotopic. For $i=1, \dots, k$ we denote
$$W_i\, =\, \{(g,h)\in G\times G; \, \, g\cdot h^{-1}\in U_i\}.$$
It is clear that $W_1\cup \dots \cup W_k$ is an open cover of $G\times G$. 
Let $h_i: U_i\times I\to G$ be a continuous homotopy, where
$I=[0,1]$, such that $h_i(x,0)=x$ and $h_i(x,1)=e$ for all $x\in U_i$, where $e\in G$ denotes the unit of $G$.
Then we may define $s_i: W_i\to PG$ by the formula
\begin{eqnarray}
s_i(A,B)(t) \, =\, h_i(A\cdot B^{-1},t)\cdot B \in G, \quad (A,B)\in W_i.
\end{eqnarray}
It is a continuous motion planning over $W_i$. This proves that $\tc(G)\leq \cat(G)$ and hence (\ref{cat}) follows
from (\ref{between}).
\end{proof}

\begin{proof}[Proof of Theorem \ref{body}]: We have to show that $\tc(SE(3))=4$; the statement will then follow from Theorem 
\ref{coincide}. 

By Lemma \ref{shmuel}, 
$$\tc(SE(3))\, =\, \cat(SE(3)).$$ 
Hence, it is enough to show that the Lusternik - Schnirelman 
category of $SE(3)$ equals 4.

$SE(3)$ is homotopy equivalent to $SO(3)\subset SE(3)$ (the subgroup of rotations). Since
the topological complexity $\tc(X)$ is homotopy invariant of $X$ (see Theorem 3 of \cite{F1}), we find
$\cat(SE(3)) = \cat(SO(3))$. On the other hand, it is well known that the special orthogonal group 
$SO(3)$ is diffeomorphic to the 3-dimensional 
projective space ${\mathbf {RP}}^3$ (the variety of all lines through the origin in $\R^4$). 
The Lusternik - Schnirelman category of any projective space ${\mathbf {RP}}^n$ equals $n+1$, see \cite{DNF}.
This completes the proof. \end{proof}

\section{Robot Arm}

\begin{figure}[h]
\begin{center}
\includegraphics[0,0][216,128]{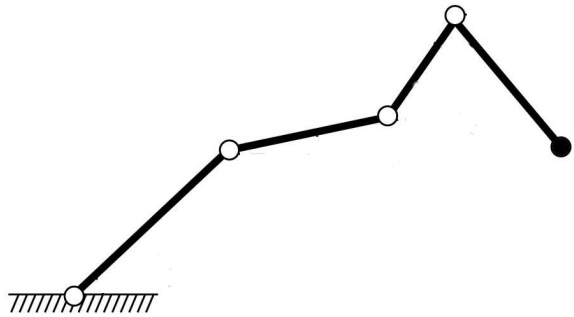}
\end{center}
\caption{Robot arm in $\R^3$.}\label{figarm}
\end{figure}

Consider a robot arm in $\R^3$ (see Figure \ref{figarm})
consisting of $n$ bars $L_1, \dots, L_n$, such that $L_i$ and $L_{i+1}$ are connected by
a flexible joint. We assume that the initial point of $L_1$ is fixed. 

The configuration space of a robot arm in the 3-dimensional space $\R^3$ is the Cartesian product of $n$ copies of the 2-dimensional sphere $S^2$:
$$X=S^2\times S^2\times \dots\times S^2= {(S^2)}^{\times n}$$ 
($n$ factors), where the factor $i$ describes the orientation in the 3-dimensional space of the bar $L_i$.

\begin{theorem}\label{arm} The topological complexity of the motion planning problem 
of an $n$-bar robot arm in $\R^3$
equals $2n+1$, i.e. 
$$\tc((S^2)^{\times n})=2n+1.$$  
Hence, any motion planner controlling a robot arm with $n$ bars,
will have degree of instability at least $2n+1$.
There exists a motion planner for a robot arm with $n$ bars, having degree of instability precisely $2n+1$. 
\end{theorem}

It is not difficult to explicitly construct motion planners for the robot arms, which have
the minimal possible topological complexity.

Theorem \ref{arm} follows from Theorem \ref{coincide} above and from Theorem 12 of \cite{F1}.

Note that for the planar robot arm with $n$ bars the configuration space is the product of $n$ circles
$$T^n=S^1\times S^1\times \dots \times S^1,$$
the $n$-dimensional torus. $T^n$ is a Lie group and hence Lemma \ref{shmuel} applies and gives $\tc(T^n)=\cat(T^n)$.
It is well-known that the Lusternik - Schnirelman category of the torus $T^n$ equals $n+1$. Hence, we find that 
$$\tc(T^n)=n+1.$$ 
This shows that for the planar robot arm with $n$ bars the minimal order of instability equals $n+1$.

\section{Avoiding Obstacles in $\R^3$}

In this section we will consider the following motion planning problem. 
Let $D_1, \dots, D_n\subset \R^3$ be a set of mutually disjoint bodies
having a piecewise smooth boundary, such that each $D_j$ is homeomorphic to the closed 3-dimensional ball
$\{x\in \R^3;| |x||\leq 1\}$. A particle, being initially in a position 
$A\in \R^3-\cup_{j=1}^n D_j$ in the complement of the
union of the bodies $D_j$, has to be moved to a final position $B\in \R^3-\cup_{j=1}^n D_j$, 
such that the movement avoids the bodies $D_1, \dots, D_n$, which represent the obstacles. 

Let us emphasize that we assume that each obstacle
$D_j$ is topologically trivial (i.e. it is homeomorphic to the ball) although we impose no assumptions on the geometrical shape
of the obstacles and on their mutual position in the space. 

The situation when the obstacles are noncompact or have a nontrivial topology will be considered
later in a separate section; we will see that the conclusions then will be slightly different.

The configuration space for this motion planning problem is the complement of the union of the bodies
$$X\, =\, \R^3-\cup_{j=1}^n D_j.$$

\begin{theorem}\label{obstacle} 
For any motion planner in the complement of the obstacles
$X =\, \R^3-\cup_{j=1}^n D_j$ there exists a pair of configurations
$(A,B)\in X\times X$
having order of instability $\geq 3$. Moreover, one may construct a motion planner
in $X$ having no pairs of initial -- final configurations $(A,B)\in X\times X$ with order of instability greater than $3$. 
\end{theorem}

\begin{proof} We may apply Theorem \ref{coincide} and hence our task is to show that $\tc(X)=3$.
From Lemma \ref{remove} below it follows that $X$ has homotopy type of a bouquet of $n$ two-dimensional spheres
$X\simeq Y_n$, where 
\begin{eqnarray}
Y_n=\,  \underbrace{S^2\vee S^2\vee\dots\vee S^2}_{n \, \, \mbox{times}}
\end{eqnarray}
denotes the bouquet of $n$ spheres $S^2$. 
Recall that a bouquet of two path-connected topological spaces is obtained from a disjoint union of these spaces
by identifying a single point in one of them with a single point in the other.
We may find a large ball $B=\{x\in \R^3;||x||\leq R\}$ with large radius $R$ which contains all the obstacles 
$D_1, \dots, D_n$ in its interior. The complement $\R^3-\cup_{j=1}^n D_j$ is homotopy equivalent to 
$B-\cup_{j=1}^n D_j$ since one may construct a (radial) deformation retraction of 
the complement of the ball $\R^3 -B$ onto the boundary $\partial B$. Now we may apply Lemma \ref{remove} several times
to obtain a homotopy equivalence $X\simeq Y_n$.

Using homotopy invariance of the topological complexity $\tc(X)$ 
(see Theorem 3 in \cite{F1}), we get $\tc(X) =\tc(Y_n)$.

Finally, we apply Lemma \ref{bouquet} below to conclude that $\tc(Y_n)=3$. 
\end{proof}
\begin{lemma}\label{bouquet}
Let $Z$ denote the bouquet of $n$ spheres $S^m$,
$$Z=\,  \underbrace{S^m\vee S^m\vee\dots\vee S^m}_{n \, \, \mbox{times}}.$$
Then 
\begin{eqnarray}
\tc(Z) =\left\{
\begin{array}{ll}
2, &\mbox{if $n=1$ and $m$ is odd},\\ \\
3 &\mbox{if either $n>1$, or $m$ is even}.
\end{array}
\right.
\end{eqnarray}
\end{lemma}
\begin{proof} 
The bouquet $Z$ is $m$-dimensional and $(m-1)$-connected. 
Therefore applying Theorem \ref{upper2} we find
$$\tc(Z) \, <\, \frac{2m+1}{m}+1\, =\, 3+\frac{1}{m}.$$
We obtain from this $\tc(Z)\leq 3$.

We want to apply Theorem \ref{lower} to obtain a lower bound for $\tc(Z)$. 
The cohomology algebra $H^\ast(Z;\R)$ has $n$ generators $u_1, u_2, \dots, u_n\in H^m(Z;\R)$ which satisfy the 
following relations: 
$$u_i u_j=0\quad \mbox{for any}\quad i, \, j.$$
Denote
$$\bar u_i = 1\otimes u_i - u_i\otimes 1\in H^\ast(Z;\R)\otimes H^\ast(Z;\R).$$
Then $\bar u_i$ is a zero-divisor (see \cite{F1}, section 4). We find that the product of two such zero-divisors
equals
$$\bar u_i \bar u_j = (-1)^{m+1} u_j\otimes u_i- u_i\otimes u_j.$$
We see that this product is nonzero, assuming that either $i\not= j$, or $i=j$ and $m$ is even.
Hence, assuming that either
$n>1$ or $n=1$ and $m$ is even,
we obtain from Theorem 7 of \cite{F1} the following lower bound $\tc(Z)\geq 3$. 

The lower and upper bounds coincide, and therefore we conclude that $\tc(Z)=3$ if either $n>1$ or
$n=1$ and $m$ is even. 

The remaining case (when $n=1$ and $m$ is odd) reduces to a single odd-dimensional sphere $S^m$; our claim 
now follows from Theorem 8 of \cite{F1}.

\end{proof}

Theorem \ref{obstacle} gives:

\begin{corollary}
The topological complexity of the motion planning problem 
in the 3-dimensional Euclidean space $\R^3$ in the presence of a number of topologically trivial obstacles
$D_1, \dots, D_n\subset \R^3$, where $n\geq 1$, does not depend on the number of obstacles and on their geometry.
\end{corollary}

\begin{example} Here we will describe an explicit motion planner with three local rules 
for the problem of moving a point in $\R^3$
avoiding the obstacles which we will represent as points $p_1, p_2, \dots, p_n$. Thus our configuration space is
$$X=\R^3 -\{p_1, p_2, \dots, p_n\}.$$
This situation may be considered as a degenerated version of the previous discussion, although topologically it is equivalent to it.

We will define explicitly three subsets $F_1, F_2, F_3\subset X\times X$ such that they are ENRs and 
form a partition of $X\times X$. Moreover, we will specify continuous maps $s_j: F_j\to PX$, where $j=1, 2, 3$,
such that $E\circ s_j=1$. 

For $F_1$ we will take the set of all pairs $(A,B)\in X\times X$, such that the Euclidean segment connecting $A$ and $B$ 
does not intersect the set of obstacles $\{p_1, p_2, \dots, p_n\}$. We will define $s_1(A,B)$ as the path which goes along the
straight line segment connecting $A$ and $B$, i.e. $s_1(A,B)(t)=(1-t)A+tB$, $t\in [0,1]$.

For $F_2$ we will take the set of all pairs $(A,B)\in X\times X$ such that the straight line segment $[A,B]$ contains some
points $p_{i_1}, p_{i_2}, \dots, p_{i_k}$ but this segment is not parallel to the $z$-axis. Our motion planning strategy $s_2(A,B)(t)$
will be to follow the path shown in Figure \ref{figround}, i.e. we move along the straight line segment $[A,B]$ until the distance to
one of the obstacles $p_{i_r}$ becomes $\epsilon$, then we move along the upper semicircle of radius $\epsilon>0$ with the center at $p_{i_r}$, 
lying in the 2-dimensional plane $P$. 
The plane $P$
contains the points $A,B$ and is parallel to the $z$-axis. Here $\epsilon >0$ is a fixed small constant
such that $||p_i-p_j||>\epsilon$ for $i\not= j$.

The set $F_3$ will consist of all $(A,B)\in X\times X$ such that the segment $[A,B]$ is parallel to the $z$-axis and 
contains some points from the set of obstacles $\{p_1, \dots, p_k\}$. The motion planning strategy $s_3$ will be similar to $s_2$ (see above) but for the plane $P$ 
we will take the plane containing $A, B$ and parallel to the $x$-axis. We pick the semicircles in the direction of the $x$-axis.

\begin{figure}[h]
\begin{center}
\includegraphics[-5,0][250, 210]{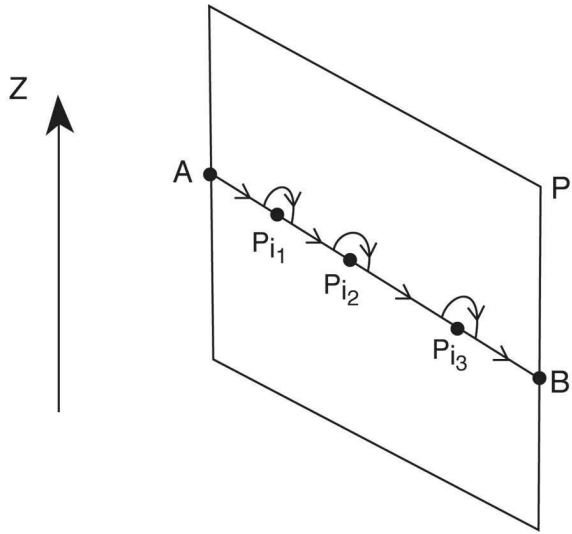}
\end{center}
\caption{Motion planning strategy avoiding obstacles}\label{figround}
\end{figure}

\end{example}

\begin{lemma}\label{remove} Let $M$ be a connected $n$-dimensional smooth 
manifold having a non-empty boundary $\partial M$.
Let $D\subset M$ be a subset homeomorphic to an $n$-dimensional ball $\{x\in \R^n; |x|\leq 1\}$,
lying in the interior of $M$ and such that the boundary $\partial D$ is piecewise smooth. Then the complement
$M-D$ is homotopy equivalent to the bouquet $M\vee S^{n-1}$.
\end{lemma}
\begin{proof}
We may find a smooth path $\gamma$ connecting a smooth point of $\partial D$ with a point of $\partial M$ (see Figure \ref{figremove}, a).
Thickening $\gamma$ we obtain a tube $T$ connecting $\partial D$ with $\partial M$, and $D\cup T$ homeomorphic to a disk
(see Figure \ref{figremove}, b). 
We see that $M- {\rm int} (D\cup T)$ is homeomorphic to $M$, since it is obtained from $M$ by 
a collapse from the boundary. 

Therefore $M- {\rm int} D$ is homeomorphic to the result of glueing the tube $T=D^2\times [0,1]$ to the manifold $M$ along $S^1\times [0,1]$.
We claim that the identification map $\phi: S^1\times [0,1] \to M$ is homotopically trivial.
To prove this it is enough to show that the small loop $\ell$ around $\gamma$ (which is the core of the image $\phi(S^1\times [0,1])$)
bounds a 2-dimensional disk in $M - {\rm int}(D\cup T)$; such a disk $\Sigma$ is shown in Figure \ref{figremove},c.

\begin{figure}[h]
\begin{center}
\includegraphics[-5,0][340, 100]{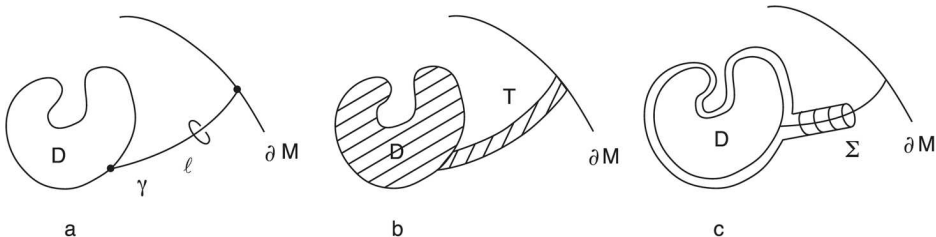}
\end{center}
\caption{Obstacle homeomorphic to a disk}\label{figremove}
\end{figure}

We obtain that $M-{\rm int D}$ is homotopy equivalent to the result of glueing to $M$ a 2-dimensional cell $D^2$
along a homotopically 
trivial map $\partial D^2 \to M$.
Hence $M-D$, which is homotopy equivalent to $M-{\rm int D}$, has homotopy type of $M\vee S^2$. 
\end{proof}

\section{Obstacles with Nontrivial Topology}

The results of the previous section become false if the obstacles are non-compact or if they have a nontrivial topology.
However the topological complexity $\tc(X)$ cannot be too large:
\begin{theorem}\label{polyhedron1}
Let $A\subset \R^3$ be a closed polyhedral subset (the obstacles) and let $X=\R^3-A$ be the complement.
Then there always exists a motion planner in $X$ with degree of instability at most $5$, i.e. $\tc(X)\leq 5$. 
\end{theorem}
\begin{proof}
$X$ is a smooth manifold and so Theorem \ref{coincide} applies. We have to show that $\tc(X)\leq 5$. We observe that $X$
is 3-dimensional but it is an open manifold (noncompact with no boundary) and thus $X$
has homotopy type of a polyhedron $Y$ of dimension 2. We know that the topological complexity
 is homotopy invariant, $\tc(X)=\tc(Y)$. Now we may apply (\ref{upper}) to obtain $\tc(X)=\tc(Y) \leq 5$. 
\end{proof}

\begin{figure}[h]
\begin{center}
\includegraphics[-10,0][350, 160]{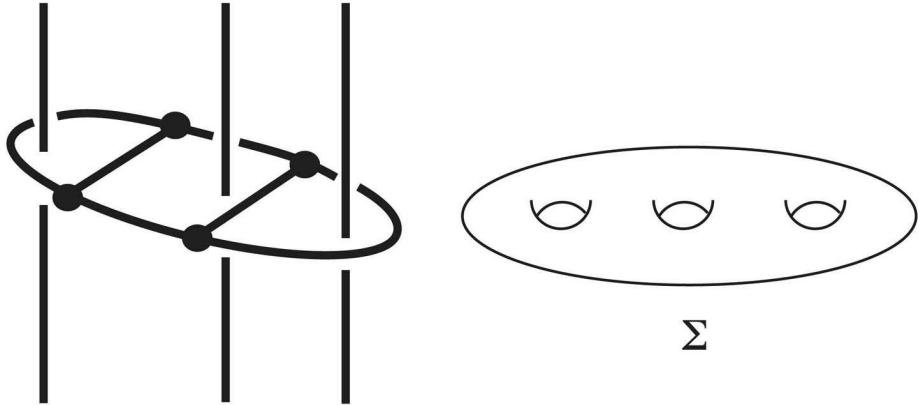}
\end{center}
\caption{Topologically nontrivial obstacles in $\R^3$ (left). 
The complement is homotopy equivalent to surface $\Sigma$ of genus 3 (right).}\label{figpipes}
\end{figure}

Consider the following example. The set of obstacles $A\subset \R^3$ is the union of two infinite tubes and a solid torus, 
see Figure \ref{figpipes}. The complement $X=\R^3 - A$ serves as a configuration space for the motion planning problem.
It is easy to see that $X$ is homotopy equivalent to a compact orientable surface $\Sigma$ of genus 3. 
Using homotopy invariance we obtain $\tc(X)=\tc(\Sigma)$. By Theorem 9 of \cite{F1} we find $\tc(\Sigma)=5$.
Therefore in this example $\tc(X) =5$. This shows that the upper bound 5 in Theorem \ref{polyhedron1} cannot be improved.

\section{Simultaneous Control of Several Systems}

Suppose that we have two different mechanical systems, which are completely independent, and our problem consists of
constructing a simultaneous motion planning program for them. 
This means that we are going to control both systems at the same time trying to bring them to a pair of desired states,
starting from a pair of initial states.

Let $X$ denote the configuration space of the first system and let $Y$ be the configuration space of the second system.
Then the configuration space, which describes the problem of simultaneous control, is $X\times Y$, the Cartesian product
of $X$ and $Y$. 

Our problem is to construct a motion planner in $X\times Y$.
It is clear that we may do so as follows.
Let $X\times X =F_1\cup F_2\cup \dots \cup F_k$, $s_j:F_j\to PX$ be a 
motion planner in $X$ and let 
$Y\times Y =G_1\cup G_2\cup \dots \cup G_\ell$, $\sigma_i:G_i\to PY$ be a motion planner in $Y$. Then 
the sets $F_j \times G_i$ give a splitting of $(X\times Y)\times (X\times Y)$ and the maps $s_j\times \sigma_i$ determine
the continuous motion planning strategies. This shows that there exists a motion planner with $k\cdot \ell$ local rules.

This straightforward approach is not optimal as the following Theorem shows:

\begin{theorem}\label{simulta}
For any path-connected metric spaces $X$ and $Y$,
\begin{eqnarray}
\tc(X\times Y) < \tc(X) +\tc(Y).
\end{eqnarray}
In other words, the topological complexity of the motion planning problem of simultaneous control of two systems
is less than the sum of their individual topological complexities.
\end{theorem}

Thus, in the situation when we have to simultaneously control several systems,
the topological complexity is at most additive and not multiplicative, as may be expected at first glance.

A proof of Theorem \ref{simulta} can be found in \cite{F1}. 

We will give here a simple explicit construction of a motion planner in $X\times Y$ with $k+\ell -1$ local rules, under an additional
assumption. This additional assumption is such that it may really be achieved in most cases. For example, the motion planner
(\ref{fine}) constructed in the proof of Theorem \ref{coincide} (when the configuration space $X$ is a manifold) 
has this property.

Assume that the motion planner $X\times X =F_1\cup F_2\cup \dots \cup F_k$, $s_j:F_j\to PX$, satisfies the following
condition: the closure of each set $F_j$ is contained in the union $F_1\cup F_2\cup \dots\cup F_j$. In other words, we require
that all sets of the form $F_1\cup F_2\cup \dots \cup F_j$ be closed. 

Similarly, we will assume that
$Y\times Y =G_1\cup G_2\cup \dots \cup G_\ell$, $\sigma_i:G_i\to PY$ is a motion planner in $Y$ such that 
all sets of the form $G_1\cup G_2\cup \dots\cup G_i$ are closed. 

Then we will set 
\begin{eqnarray}
W_r = \bigcup_{j+i=r} F_j\times G_i, \quad r=2, 3, \dots, k+\ell.\label{prd}
\end{eqnarray}
The sets are ENRs and form a partition of $(X\times X)\times (Y\times Y) = (X\times Y)\times (X\times Y)$.
Our assumptions guarantee that each product $F_j\times G_i$ is closed in $W_r$, where $r=j+i$.
Since different products in the union (\ref{prd}) are disjoint, we see that the maps $s_j\times \sigma_i$, where $j+i=r$, 
determine a continuous motion planning strategy over each set $W_r$.

\begin{example} Let $A\subset \R^3$ be the set of obstacles shown in Figure \ref{figpipes}. Consider the problem of simultaneous control
of $n$ independent particles lying in the complement $X=\R^3- A$. The configuration space 
$$Y \, =\, X^{\times n}=X\times X\times \dots \times X$$
is the Cartesian product of $n$ copies of $X$. We claim that
$$\tc(Y) \, =\, \tc(X^{\times n}) \, =\,4n+1,$$
and hence:  (1) {\it any motion planner for the problem will have order of instability $\geq 4n+1$ and at least $4n+1$ local rules,} 
and (2) {\it there exists
a motion planner with order of instability $4n+1$ having precisely $4n+1$ local rules. }

Since $X$ is homotopy equivalent to the surface $\Sigma$ of genus 3, we obtain that $\tc(X^{\times n}) = \tc(\Sigma^{\times n})$.
By Theorem 9 of \cite{F1}, $\tc(\Sigma)=5$, and hence, applying inductively Theorem \ref{simulta} we obtain an inequality
$\tc(\Sigma^{\times n})\leq 4n+1$. 

To find a lower bound for $\tc(\Sigma^{\times n})$ we will apply Theorem \ref{lower}. Let 
$$a,b,c,d\in H^1(\Sigma;\R)$$
be a symplectic basis 
$$a^2=b^2=c^2=d^2 =0, \quad ab =A, \quad cd =A,$$
where $A\in H^2(\Sigma;\R)$ is a fundamental class. We may also assume that 
$$ac=ad=bc=bd=0.$$
For $i=1, 2, \dots, n$ denote 
$$a_i=1\times \dots\times  1\times  a\times 1\times \dots \times 1\in H^1(\Sigma^{\times n};\R).$$
Here the class $a$ appears in place $i$. We will define similarly the cohomology classes
$$b_i, c_i, d_i\in H^1(\Sigma^{\times n};\R), \quad i=1, \dots, n.$$
The class
$$\bar a_i =1\otimes a_i-a_i\otimes 1\, \in\,  H^\ast(\Sigma^{\times n};\R)\otimes H^\ast(\Sigma^{\times n};\R)$$
belongs to the ideal of the zero-divisors. Similarly, we will define the classes
$$\bar b_i, \, \bar c_i, \, \bar d_i \, \in\,  H^\ast(\Sigma^{\times n};\R)\otimes H^\ast(\Sigma^{\times n};\R), \quad i=1, \dots, n,$$
lying in the ideal of the zero-divisors. 

We claim that the product 
\begin{eqnarray}
\prod_{i=1}^n (\bar a_i\bar b_i\bar c_i\bar d_i) \, \not= \, 0\label{cohom}
\end{eqnarray}
is nonzero. We compute
$$\bar a_i \bar b_i = 1\otimes A_i +b_i\otimes a_i-a_i\otimes b_i -A_i\otimes 1,$$
where $A_i=1\times \dots 1\times A\times 1\times \dots\times 1\in H^2(\Sigma^{\times n};\R)$; the cohomology class $A$ appears in place $i$.
Similarly we find
$$\bar c_i \bar d_i = 1\otimes A_i +d_i\otimes c_i-c_i\otimes d_i -A_i\otimes 1,$$
and therefore
$$\bar a_i \bar b_i\bar c_i \bar d_i = -2 A_i\otimes A_i.$$
Hence, we see that product (\ref{cohom}) equals 
$$(-2)^n \cdot U\otimes U\not=0,$$
where $U=A\times A\times \dots \times A\in H^{2n}(\Sigma^{\times n};\R)$. Now, Theorem \ref{lower} applies and gives
$$\tc(\Sigma^{\times n}) \geq 4n+1.$$
Indeed, product (\ref{cohom}) contains $4n$ factors which are all zero-divisors.

This proves that in this motion planning problem, when we have to simultaneously control $n$ 
independent particles, the topological complexity equals $4n+1$, in particular it is 
a linear function of the number of particles.
\end{example}
\vskip 1cm

{\bf Acknowledgement}. I am gratefulful to D. Halperin, M. Sharir, S. Tabachnikov and S. Weinberger for useful discussions.

\bibliographystyle{amsalpha}

\begin{thebibliography}{999}

\bibitem{D} A. Dold, \textit{Lectures on Algebraic Topology}, Springer - Verlag, 1972.

\bibitem{DNF} B. Dubrovin, S. P. Novikov and A. T. Fomenko, \textit{Modern Geometry; Methods of
the Homology Theory}, 1984.

\bibitem{F1} M. Farber, \textit{Topological complexity of motion planning}, Preprint math.AT/0111197


\bibitem{FHS} M. Farber, D. Halperin, M. Sharir, \textit{Continuous motion planners and topological complexity of configuration spaces}, Technical report, 2002.

\bibitem{J} I.M. James, \textit{On category, in the sense of Lusternik - Schnirelman}, Topology, \textbf{17}(1978), 331 - 348.

\bibitem{K} J. Kelley, \textit{General Topology}, New York, 1957.

\bibitem{L} J.-C. Latombe, \textit{Robot Motion Planning}, Kluwer Academic Publishers, 1991.

\bibitem {Sz} A. S. Schwarz, \textit{The genus of a fiber space}, Amer. Math. Sci. Transl.
\textbf{55}(1966), 49 - 140.

\bibitem{Sh} M. Sharir, \textit{Algorithmic motion planning}, Handbook of 
Discrete and Computational Geometry,
J. Goldman, J. O'Rourke, editors, 1997, CRC Press.

\bibitem{Sm1} S. Smale, \textit{On the topology of algorithms, I}, 
J. of Complexity, \textbf{3}(1987), 81 - 89.

\bibitem{Sp} E. Spanier, \textit{Algebraic Topology}, 1966.

\bibitem{V1} V.A. Vassiliev, \textit{Cohomology of braid groups and complexity of algorithms},
Functional Analysis and its Appl., \textbf{22}(1988), 15 - 24.











\end{thebibliography}

\end{document}